\documentclass[conference]{IEEEtran}
\usepackage[latin9]{inputenc}
\usepackage{amsmath, amssymb, amsfonts}
\usepackage{graphicx}
\usepackage[caption=false, font=footnotesize]{subfig}
\usepackage{stfloats}
\usepackage{url}
\usepackage{glossaries}
\usepackage{diagbox}
\usepackage{balance}
\usepackage{xcolor}
\usepackage{booktabs}
\usepackage{makecell}
\usepackage{tabularx,ragged2e}
\newcolumntype{?}{!{\vrule width 2\arrayrulewidth}}
\usepackage{xspace} 
\usepackage{comment}

\definecolor{mat_orange}{HTML}{FF7F0E}
\definecolor{mat_blue}{HTML}{1F77B4}
\definecolor{yellow}{rgb}{0.93, 0.69, 0.13}

\newacronym{roc}{ROC}{Receiver Operating Characteristic}
\newacronym{auc}{AUC}{Area Under the Curve}
\newacronym{snr}{SNR}{Signal-to-Noise Ratio}
\newacronym{mse}{MSE}{Mean Squared Error}
\newacronym{fdr}{FDR}{Free Decay Region}
\newacronym{rir}{RIR}{Room Impulse Response}
\newacronym{rt}{RT}{Reverberation Time}
\newacronym{stft}{STFT}{Short Time Fourier Transform}
\newacronym{dft}{DFT}{Discrete Fourier Transform}
\newacronym{cnn}{CNN}{Convolutional Neural Network}
\newacronym{dct}{DCT}{Discrete Cosine Transform}
\newacronym{svm}{SVM}{Support Vector Machine}
\newacronym{gan}{GAN}{Generative Adversarial Network}

\begin{document}
%
\title{Training CNNs in Presence of JPEG Compression:\\ Multimedia Forensics vs Computer Vision}

\author{
\IEEEauthorblockN{Sara Mandelli, Nicol\`o Bonettini, Paolo Bestagini, Stefano Tubaro}
\IEEEauthorblockA{Dipartimento di Elettronica, Informazione e Bioingegneria\\ Politecnico di Milano, Piazza Leonardo da Vinci 32, 20133 Milano, Italy\\
}
}

\maketitle

\begin{figure}[b]
\vspace{-0.3cm}
\parbox{\hsize}{\em
WIFS`2020, December, 6-9, 2020, New York, USA.
XXX-X-XXXX-XXXX-X/XX/\$XX.00 \ \copyright 2020 IEEE.
}\end{figure}

\begin{abstract}
\glspl{cnn} have proved very accurate in multiple computer vision image classification tasks that required visual inspection in the past (e.g., object recognition, face detection, etc.).
Motivated by these astonishing results, researchers have also started using \glspl{cnn} to cope with image forensic problems (e.g., camera model identification, tampering detection, etc.).
However, in computer vision, image classification methods typically rely on visual cues easily detectable by human eyes.
Conversely, forensic solutions rely on almost invisible traces that are often very subtle and lie in the fine details of the image under analysis.
For this reason, training a \gls{cnn} to solve a forensic task requires some special care, as common processing operations (e.g., resampling, compression, etc.) can strongly hinder forensic traces.
In this work, we focus on the effect that JPEG has on \gls{cnn} training considering different computer vision and forensic image classification problems.
Specifically, we consider the issues that rise from JPEG compression and misalignment of the JPEG grid.
We show that it is necessary to consider these effects when generating a training dataset in order to properly train a forensic detector not losing generalization capability, whereas it is almost possible to ignore these effects for computer vision tasks.
\end{abstract}

\IEEEpeerreviewmaketitle

\section{Introduction}
\label{sec:intro}

Thanks to the increasing availability of digital data and computational resources, \glspl{cnn} have greatly outperformed multiple classical approaches for a wide variety of tasks in different fields \cite{Wang2018, Reichstein2019}.
For instance, they have been used with outstanding results to solve several computer vision problems related to image analysis and understanding.
This is the case of object detection \cite{Ren2015}, image segmentation \cite{Ronneberger2015}, image retrieval \cite{Yue2015} and many other tasks.
All these tasks are characterized by two properties:
(i) they are extremely challenging to solve using a model-based solution (i.e., it is hard to define objective properties that an image must have in order to belong to a given class);
(ii) they can be reasonably well solved by human operators through  visual inspection (i.e., a human can classify or segment an object in a scene using visual semantic cues, given that the quality of the image under analysis is decent).
In other words, we often know what we are looking for, but we cannot describe it easily.

The rise of \gls{cnn}-based solutions over classical methods is not just a computer vision prerogative.
In the last few years, also the multimedia forensics community has started replacing more classical detectors with \gls{cnn}-based ones.
Camera model identification \cite{tuama2016, bondi2016}, image tampering localization \cite{Bondi2017, cozzolino2020} and deepfake detection \cite{Guera2018, bonettini2020} are just a few examples of problems whose accurate solutions are nowadays based on the use of \glspl{cnn}.
However, despite forensic and computer vision tasks share some similarities, they also have some differences.
On one hand, forensic problems are often hard to solve through purely model-based methods as computer vision tasks.
As an example, considering an image tampering detection problem, defining a complete model accommodating for all possible image forgery operations is far from being practical.
On the other hand, forensic tasks cannot be typically solved using visual cues by human operators.
As an example, it is almost impossible to tell how many compression steps an image underwent or which is the specific camera model used for the shooting, even if the image quality is extremely high.

The impossibility of easily extracting forensic information by simple visual inspection is due to the fact that forensic traces are often hidden in tiny details of the image under analysis.
For instance, forensic information can lie in high-pass components, or in low-power and noisy-like signals hardly detected by human eyes.
For this reason, training a \gls{cnn} for a  computer vision task or for a forensic one can be strongly different.
As an example, we expect that typical image processing and enhancement operations do not impact on object detection tasks (unless operated as an adversarial attack \cite{Huang2017}).
In other words, the class of an image should not change even if the image is resized, compressed, undergone some color correction or other operations, as far as a human operator can still recognize the object.
Conversely, it is hard to tell whether processing operations impact on forensic traces of any kind, as we do not always clearly understand which are the important cues that a \gls{cnn} captures.

In this paper, we specifically focus on problems introduced by JPEG compression while training a \gls{cnn} for forensic tasks.
Indeed, JPEG is one of the most widely used image compression standard and is well-known to hinder forensic traces.
We analyze the effect of working with JPEG compressed images which have been cropped with respect to their original size and the consequence of applying JPEG compression with different quality factors during training and/or testing of a \gls{cnn}.
We compare results achieved on two forensic tasks (i.e., camera model identification and detection of synthetically generated images) with those achieved on two computer vision tasks (i.e., image classification on ImageNet \cite{Deng2009} and Lsun \cite{Yu2015} datasets).
For each task, we use four different \glspl{cnn} in order to avoid possible biases due to some specific architectures.
Results confirm that training a \gls{cnn} for a forensic task needs some special care that is not necessary for computer vision tasks.

We hope this analysis can be helpful to practitioners and researchers willing to use powerful data-driven approaches in forensic scenarios, as they might risk getting tricked in learning something not related to the task under analysis.
We also hope our experiments will motivate some more research in this area.


\section{Background}
\label{sec:background}

\subsection{JPEG Compression Forensics}
JPEG is one of the most widely used compression scheme on the web, and it approximately works as follows.
Input images are partitioned into $8 \times 8$ non-overlapped pixel blocks. 2D \gls{dct} is computed for each block and  transform coefficients are quantized into integer-valued levels according to the selected quantization matrix and quality factor. Quantization is the step causing information loss. A lower quantization quality factor indicates a stronger quantization, thus lower quality of the final decompressed image. Quantized values are then converted into a binary stream by means of lossless coding. At decoding time, the binary stream is decompressed, coded blocks are reconstructed by applying inverse 2D \gls{dct} on quantized coefficients and the image is re-built in the pixel domain.

JPEG compression leaves peculiar traces which have been studied in the forensics literature for many years.
For instance, the $8 \times 8$ block processing leaves characteristic blocking artifacts that can be exploited to tell whether an image underwent JPEG compression \cite{fan2003}.
It is also well known that quantization affects \gls{dct} coefficients' histograms, influencing in peculiar way their shape depending on the number of compressions an image has undergone.
This can be exploited to detect aligned and non-aligned JPEG compressions \cite{farid2009, bianchi2012, barni2010}.
Alternatively, the authors of \cite{Milani2014, pasquini2014} propose methods based on the first digits' law to detect multiple compressions. The same task is accomplished in \cite{Mandelli2018} by means of non-negative matrix factorization, which is particularly useful given the non-negative nature of histograms. Primary quantization step in double compression is estimated by statistical model of \gls{dct} in \cite{thai2019}.
Since the JPEG compression pipeline has several degrees of freedom in its definition (e.g., the quantization rule, the quantization matrix corresponding to a certain quality factor, etc.), it is also possible to distinguish among different implementations (i.e., different software used for the compression) as shown in \cite{agarwal2017, Bonettini2018}.

\subsection{CNNs in Multimedia Forensics}
In the last few years, following a common trend in many fields, \glspl{cnn} have outperformed many classical forensics detectors. 
Camera model identification is considered in \cite{bondi2016, tuama2016, bayar2018}, where deep features are extracted from images by means of \glspl{cnn} and then fed to model based algorithms to perform the classification task.
In \cite{bonettini2017, ahn2020} authors show how it is possible to train a \gls{cnn} to detect double JPEG compression, easing its work by pre-computing \gls{dct} coefficients.
In~\cite{barni2018} authors makes use of a \gls{cnn} approach for detecting contrast adjustment on images in presence of JPEG compression. 
\glspl{cnn} have successfully outperformed classical detectors also for the source device identification problem. In \cite{kirchner2019}, a CNN-based image denoising is proposed instead of classical denoising procedures for boosting device attribution performance, while in \cite{Mandelli2020} the \gls{cnn} is forced to learn a way of comparing camera fingerprint and image noise at patch level.
Given the flourish of generation techniques such as \glspl{gan} and DeepFakes~\cite{verdoliva2020}, the forensics community has recently focused on spotting media generated in such manner.  In \cite{marra2019gan} authors show how different \glspl{gan} could leave different traces in generated images,  giving the investigator a tool to detect them. The problem of incremental learning on new kind of \glspl{gan} is considered in \cite{marra2019incremental}. A method for detecting DeepFake videos using Recurrent Neural Network is proposed in \cite{Guera2018}, while in \cite{bonettini2020} an ensemble of networks with an attention mechanism is employed for the same task. In this vein, a study on different preprocessing and augmentations techniques (including JPEG compression) when dealing with detection of \gls{cnn} generated images has been proposed in \cite{Wang2020}.


\section{Case Studies}
\label{sec:case_studies}
In order to investigate the impact of JPEG when training \glspl{cnn} for multimedia forensics and computer vision tasks,
we consider two realistic scenarios as case studies: (i) the dataset under analysis contains some JPEG images cropped with respect to their original size; (ii) the dataset contains some JPEG images that have been compressed with unknown quality factor.
It follows an exhaustive description of each scenario.

\subsection{JPEG Grid Misalignment}
While editing a photograph or simply uploading a profile picture over social networks, it often occurs that images are cropped with respect to their original size.
As a matter of fact, this operation is performed most of the times without paying attention to the precise pixel coordinates of the cropped area, as it is more important to prevent the picture subjects being canceled by the cropping.
As a consequence, it generally happens that JPEG-compressed images are cropped without respecting the $8 \times 8$ characteristic pixel grid introduced by JPEG compression. If the image is then further saved as JPEG, a new $8 \times 8$ grid non-aligned with the original one is generated.

When training a \gls{cnn} to solve an image classification problem, the presence of JPEG grid misalignment on images can cause issues depending on the specific task.
In Section~\ref{sec:experiments}, we perform some experiments showing when JPEG misalignment can be problematic considering both image forensics and computer vision tasks.

\subsection{Quality factor of JPEG Compression}
When collecting images ``in the wild", the fact that images may come from different cameras, may be post-processed or uploaded on social media multiple times has to be taken for granted. 
Likewise, it is well known that each single camera model, image editing software or social media may compress images using different JPEG implementations and/or parameters (e.g., quantization matrix, quality factors, etc.).
As a result, we can state that a wide variety of differently compressed JPEG images can be found on the internet.

Compressing images with different quality factors surely leave peculiar forensic traces.
For this reason, we aim at investigating how the quality factor of JPEG compressed images affects a \gls{cnn} performance. In Section~\ref{sec:experiments}, we analyze computer vision and image forensics tasks, eventually showing how the training phase can be tuned in order to improve results in case different JPEG compressions are used.

\section{Experimental Setup}
\label{sec:experiments}

\subsection{Datasets}
In order to provide a sufficiently general idea of the impact of JPEG compression on multimedia forensics and computer vision problems, we consider two tasks per area.

\vspace{.5em}\noindent\textbf{Multimedia Forensics.}
Regarding multimedia forensics, we tackle two problems: (i) camera model identification, i.e., identifying the source camera model of a query image; (ii) detection of \gls{cnn}-generated images from original photographs.
If the first problem is common in forensics investigation, the latter is taking the trend in the last few years, due to the widespread diffusion of images and videos with fake content produced by means of \gls{cnn}-based technologies.
To investigate these problems, we consider well known datasets available in the literature.
For camera model identification task, we exploit images selected from the Vision dataset \cite{Shullani2017}, considering the $28$ available camera models. For each camera model, we pick all the natural images (more than $150$ images per model), extracting $10$ patches per image with a common size of $512 \times 512$ pixels. 
Concerning the task of CNN-generated images, we select images from the dataset released in \cite{marra2019gan}, which provides pairs of generated/pristine images for multiple CNN-generated image categories (e.g., apple2orange, summer2winter, etc.). Specifically, we select images generated from CycleGAN \cite{zhu2017} ($\sim 1000$ images per category), extracting one patch per image and cropping to $224 \times 224$ pixels.

\vspace{.5em}\noindent\textbf{Computer Vision.}
As extremely common computer vision task, we considered image object recognition, i.e., learning to classify images according to the specific subject depicted. We investigate image classification performances of CNNs over two datasets.
Images of the first dataset are selected from $16$ different synsets of the well known ImageNet database \cite{Deng2009}. The second dataset is extracted from Lsun dataset \cite{Yu2015}, where images from $20$ different categories are selected. 
In both cases, we extract one patch per image, cropping to a common resolution of $224 \times 224 $ pixels.
In detail, we pick around $1000$ and $2000$ images per class for ImageNet and Lsun, respectively.

\subsection{Network training and evaluation}
We investigate four different networks. 
Two networks are selected from the recently proposed EfficientNet family of models \cite{Tan2019}, which achieves very good results both in computer vision and multimedia forensics tasks. Specifically, we select EfficientNetB0 and EfficientNetB4 models. 
The other networks are known in literature as ResNet50~\cite{He2016} and XceptionNet~\cite{Chollet2017}.

Following a common procedure in CNN training, we initialize the network weights using those trained on ImageNet database.
All \glspl{cnn} are trained using cross-entropy loss and Adam optimizer with default parameters.
The learning rate is initialized to $0.001$ and is decreased by a factor $10$ whenever the loss does not decrease for $10$ epochs.
Training is stopped if loss does not decrease for more than $20$ epochs, and the model providing the best validation loss is selected.

Concerning the dataset split policy, we always keep $80\%$ of the images for training phase (further divided in $85\%-15\%$ for training and validation sets, respectively), leaving the remaining $20\%$ to the evaluation set.
All the experiments are performed on the evaluation set in closed-set scenario, i.e., given a query image, the correct image category is always present in the given set of possible answers. 
After the network prediction, the category returning the highest CNN score is associated with the image. 
We use the average accuracy of correct predictions per category as evaluation metrics.

All tests have been run on a workstation equipped with one Intel\textsuperscript{\scriptsize{\textregistered}}
Xeon Gold 6246 (48 Cores @3.30 GHz), RAM 252 GB, one TITAN RTX (4608 CUDA Cores @1350 MHz), 24 GB, running Ubuntu 18.04.2.
We resort to Albumentation \cite{Buslaev2020} as our data augmentation library, while we use Pytorch \cite{Paszke2019} as Deep Learning framework.

\subsection{Experiments}
The four tasks and four networks have been tested considering the impact of two different JPEG compression aspects.

\vspace{.5em}\noindent\textbf{JPEG-grid misalignment.}
In this experiment, for both training and test, we consider the combinations of two possible scenarios: (i) images are always aligned to the $8 \times 8$ JPEG grid that starts from the top-left pixel; (ii) images are non-aligned with the JPEG grid, which may start from any pixel position.
To simulate this effect, we JPEG compress all images using a quality factor of $100$. This guarantees that images contain the typical JPEG lattice, not worsening their visual quality in any way.
To obtain image patches coherently aligned to the JPEG lattice, we select patches by cropping the images in a way that patches are aligned to the $8\times 8 $ pixel grid.
To obtain patches not aligned with the grid, we extract them in random positions.


\vspace{.5em}\noindent\textbf{Quality factor of JPEG compression.}


In this experiment, we consider that evaluation images may be compressed with different quality factors $\mathrm{QF} \in \{50, 60, 70, 80, 90, 99 \}$.
We perform all possible combinations of training and testing under different hypothesis on the used quality factor.
Specifically, we train the \glspl{cnn} in two ways: (i) using image patches directly selected from the original datasets, not applying any kind of data augmentation; (ii) performing data augmentation in training phase, including in the training dataset half of the images compressed with a quality factor selected from the above reported list.
In the first training situation, we suppose to know nothing about the JPEG compression parameters of the evaluation images.
In the second scenario, we assume some knowledge on the JPEG quality factor and exploit this to potentially improve final results.
In this setup, patches are extracted in random positions, being the JPEG-grid alignment a nuisance parameters for the evaluation of accuracy versus JPEG quality factor.


\section{Results}\label{sec:results}

\begin{figure}[t]
	\centering
	\includegraphics[width=\columnwidth]{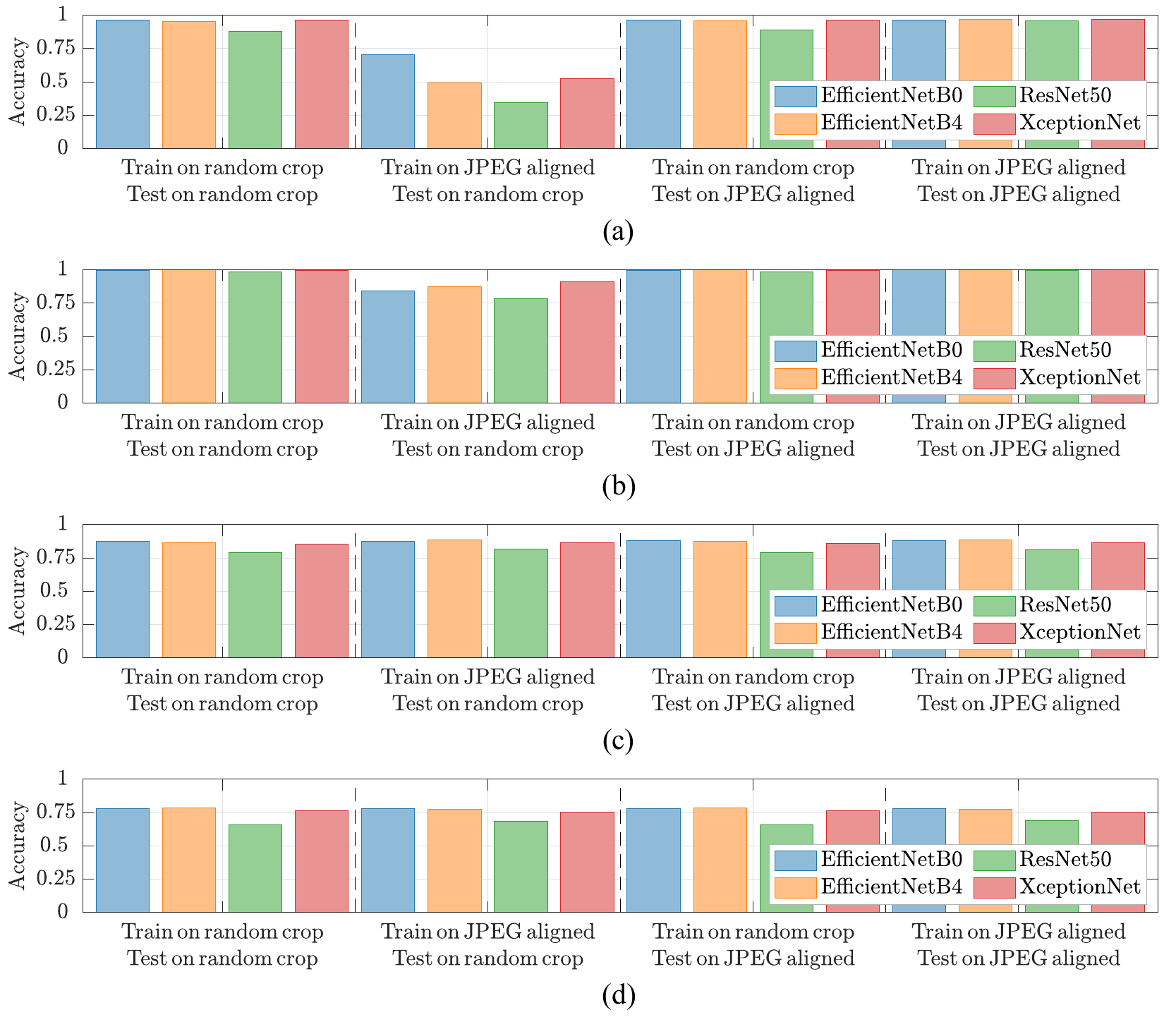}
	\caption{Accuracy as a function of JPEG-grid alignment in training and/or testing phases, for the tasks of: (a) camera model identification; (b) CNN-generated image detection; (c) image classification on ImageNet subset; (d) image classification on Lsun subset.}
	\label{fig:jpeg_misalignmentV}
\end{figure}

\vspace{.5em}\noindent\textbf{JPEG-grid misalignment.}
Fig.~\ref{fig:jpeg_misalignmentV} reports JPEG grid misalignment results for the problems of camera model identification (a), CNN-generated image detection (b) and image classification on the extracted subsets of ImageNet (c) and Lsun (d) databases.

Considering multimedia forensics tasks, we notice that being careful to the JPEG grid alignment of the extracted patches is paramount for achieving good accuracy.
When we train a detector on JPEG-aligned patches and we test it on JPEG-misaligned ones, results drop consistently.
This is especially true for the camera model identification problem (Fig.~\ref{fig:jpeg_misalignmentV}a), where all CNNs report accuracies always higher than $0.88$, except for the case of training on JPEG-aligned images and testing on random cropped ones. In this scenario, none of the proposed CNN architectures is able to overcome $0.71$ as average accuracy.
For the task of CNN-generated image detection (Fig.~\ref{fig:jpeg_misalignmentV}b), average accuracy worsens as well only in this particular situation.

On the contrary, if we consider the computer vision tasks (Fig.~\ref{fig:jpeg_misalignmentV}c and Fig.~\ref{fig:jpeg_misalignmentV}d), training and/or testing on patches aligned or not with the JPEG grid does not change the achieved results in a systematic way.
Results are almost uniform for all the networks.
This probably means that CNNs are really capturing some information related to visual cues which are not hindered by a simple JPEG-grid misalignment.

\begin{figure*}[t]
	\centering
	\includegraphics[width=.85\textwidth]{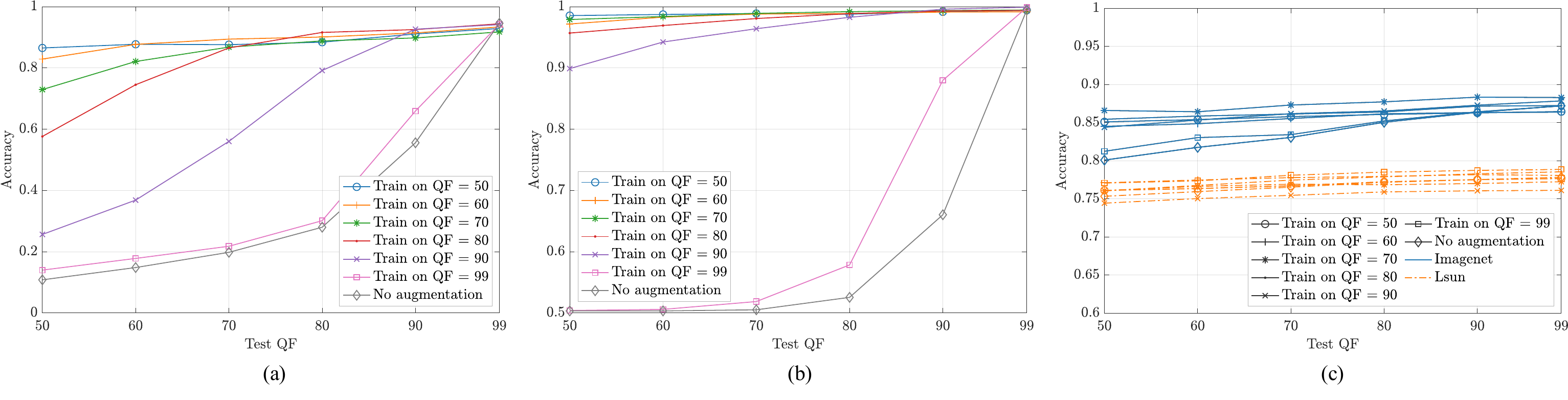}
	\caption{Accuracy of EfficientNetB0 as a function of training augmentation for the tasks of (a) camera model identification, (b) CNN-generated image detection, (c) image classification on ImageNet and Lsun subsets.}
	\label{fig:effb0_quality}
\end{figure*}

\begin{figure*}[t]
	\centering
	\includegraphics[width=.85\textwidth]{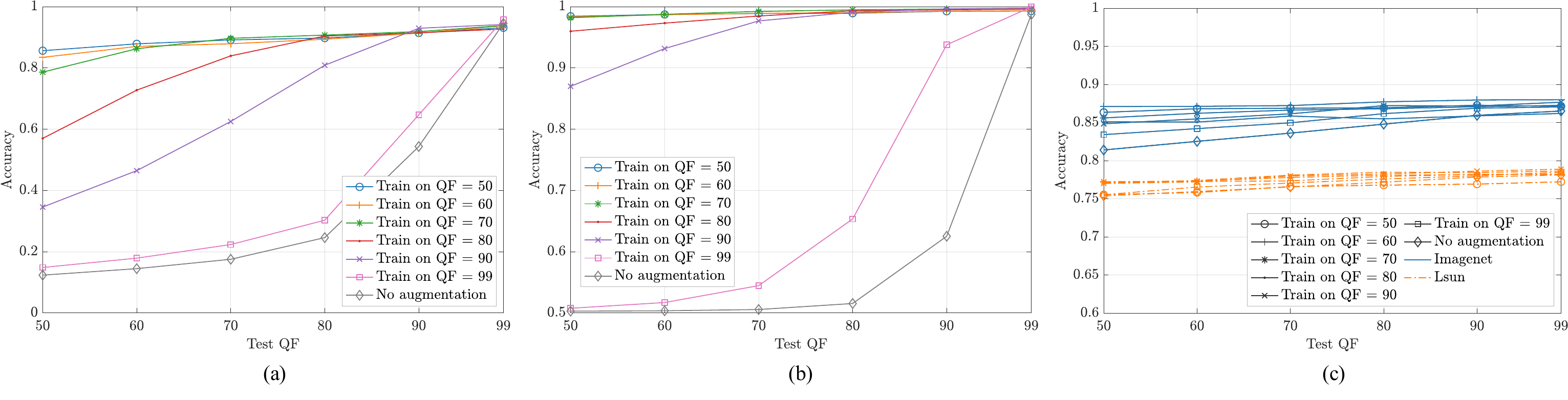}
	\caption{Accuracy of EfficientNetB4 as a function of training augmentation for the tasks of (a) camera model identification, (b) CNN-generated image detection, (c) image classification on ImageNet and Lsun subsets.}
	\label{fig:effb4_quality}
\end{figure*}

\begin{figure*}[t]
	\centering
	\includegraphics[width=.85\textwidth]{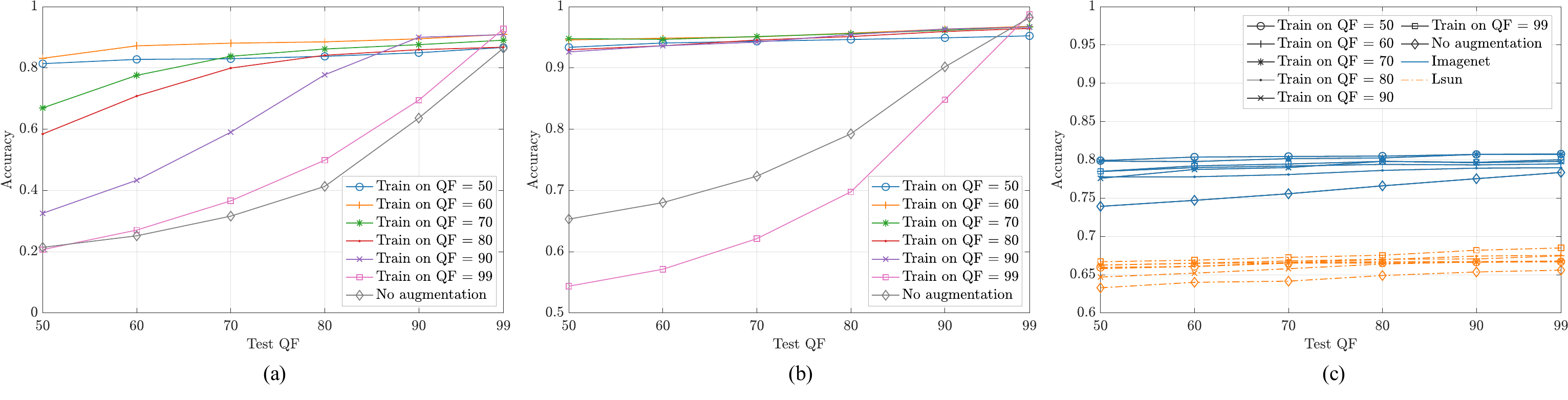}
	\caption{Accuracy of ResNet50 as a function of training augmentation for the tasks of (a) camera model identification, (b) CNN-generated image detection, (c) image classification on ImageNet and Lsun subsets.}
	\label{fig:resnet_quality}
\end{figure*}

\begin{figure*}[h!]
	\centering
	\includegraphics[width=.85\textwidth]{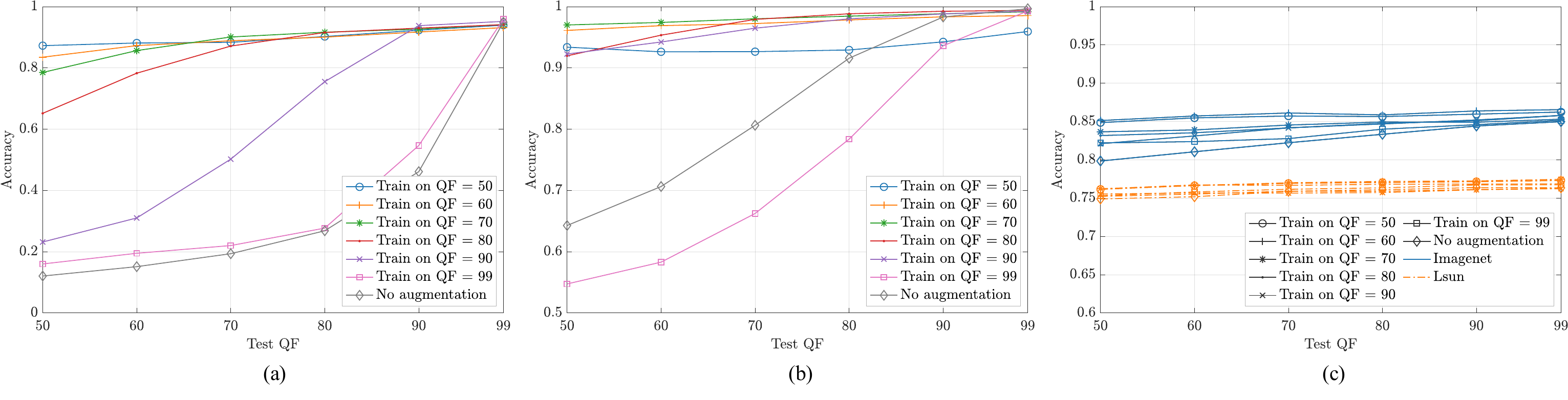}
	\caption{Accuracy of XceptionNet as a function of training augmentation for the tasks of (a) camera model identification, (b) CNN-generated image detection, (c) image classification on ImageNet and Lsun subsets.}
	\label{fig:xcep_quality}
\end{figure*}

\vspace{.5em}\noindent\textbf{Quality factor of JPEG compression.}
Figs. \ref{fig:effb0_quality}, \ref{fig:effb4_quality}, \ref{fig:resnet_quality} and \ref{fig:xcep_quality} depict results related to investigations on the JPEG quality factor for EfficientNetB0, EfficientNetB4, ResNet50 and XceptionNet, respectively.

In these experiments as well, notice that multimedia forensics tasks suffer from JPEG quality factor mismatching much more than computer vision ones. 
For instance, looking at the camera model identification goal, not performing augmentation in training phase can strongly hinder CNN performances: by training on augmented data with $\mathrm{QF} = 50$, evaluation accuracy can pass from less than $0.2$ to more than $0.85$. 
Regarding all the multimedia forensics tasks, 
selecting a low quality factor for training augmentation seems to allow a better coverage in testing phase with respect to high quality factors. 
Indeed, training on data augmented with $\mathrm{QF} = 99$ almost corresponds to absence of augmentation and achieves acceptable results only when the test $\mathrm{QF}$ matches the training one.
On the contrary, as training $\mathrm{QF}$ decreases, evaluation results present a flat behavior for all the possible test $\mathrm{QF}$. 

This phenomenon does not occur or is extremely reduced in computer vision problems, where the dynamic range of output accuracy is much more limited around the same average value in all the experiments. 
Again, this probably means that computer vision goals are not influenced by training/testing on specific JPEG configurations, and results maintain accurate whenever the image quality is preserved and objects remain detectable.
Conversely, in multimedia forensics, image visual quality is not the only thing to take care of.



\section{Conclusions}
\label{sec:conclusions}

In this work, we studied the effect of JPEG compression on \glspl{cnn} applied to multimedia forensic tasks.
In particular, we considered the effect of training and testing \glspl{cnn} considering different JPEG-grid alignment and JPEG quality factors.
We compared the achieved results to those obtained in the same conditions on two computer vision tasks.

Results show that \glspl{cnn} are extremely delicate in the multimedia forensic scenario.
If we train a \gls{cnn} only considering uncompressed images, it fails when applied to compressed ones.
If we train a \gls{cnn} only considering a specific JPEG-grid alignment, it will fail on randomly cropped images.
Conversely, computer vision tasks involving image analysis and understanding are inherently more robust to JPEG compression, given that image visual quality remains decent.

In the light of these results, we will pay particular attention whenever we train a \gls{cnn} or another sophisticated data-driven method for forensic purpose, as we want to avoid getting biased by some specific global processing traces.

\section*{Acknowledgment}
This material is based on research sponsored by DARPA and Air Force Research Laboratory (AFRL) under agreement number FA8750-16-2-0173. The U.S. Government is authorized to reproduce and distribute reprints for Governmental purposes notwithstanding any copyright notation thereon. 
The views and conclusions contained herein are those of the authors and should not be interpreted as necessarily representing the official policies or endorsements, either expressed or implied, of DARPA and Air Force Research Laboratory (AFRL) or the U.S. Government. Hardware support was generously provided by NVIDIA Corporation.

\bibliographystyle{IEEEtran}
\bibliography{biblio}

\end{document}